\documentclass[11pt, letterpaper]{article}
\usepackage{authblk}
\usepackage[margin=1in]{geometry}

\usepackage{amsmath}
\usepackage{amssymb}
\usepackage[numbers,square,comma,sort&compress]{natbib}
\usepackage{booktabs}
\newcommand{\botrule}{\bottomrule}
\usepackage{framed}
\usepackage{graphicx}  
\usepackage{xcolor}
\definecolor{myred}{HTML}{750014}
\definecolor{mygreen}{HTML}{3F7D20}
\definecolor{myochre}{HTML}{9A6D38}
\usepackage{rotating}
\usepackage{pdflscape}
\usepackage{float}
\usepackage{placeins}
\usepackage{caption}
\usepackage{makecell}
\usepackage{textcomp}
\usepackage{tikz}
\usetikzlibrary{arrows.meta}
\usepackage{booktabs}
\usepackage{hyperref}

\hypersetup{
    colorlinks=true,   
    citecolor=blue,    
    linkcolor=blue,    
    urlcolor=blue      
}

\begin{document}

\title{DySCo: Dynamically consistent data-driven downscaling of extremes in climate projections}

\author[1]{Stamatios Stamatelopoulos}
\author[1,2]{Mengze Wang}
\author[3]{Ignacio Lopez-Gomez}
\author[3]{Leonardo Zepeda-N\'u\~nez}
\author[3]{Zhong Yi Wan}
\author[3]{Robert Carver}
\author[3,{\dag}]{Fei Sha}

\author[1,*]{Themistoklis P. Sapsis}

\affil[1]{{Department of Mechanical Engineering}, 
{Massachusetts Institute of
Technology},
{Cambridge}, {MA}, {USA}}
\affil[2]{{Department of Mechanical Engineering}, 
{City University of Hong Kong}, 
{{Kowloon, Hong Kong}, {China}}}
\affil[3]{{Google Research}, {Mountain View}, {CA}, {USA}}

\maketitle

\begingroup
\renewcommand{\thefootnote}{*}
\footnotetext[1]{\sffamily\scriptsize Corresponding author: \href{email:sapsis@mit.edu}{sapsis@mit.edu}}
\renewcommand{\thefootnote}{\dag}
\footnotetext[1]{\sffamily\scriptsize Current affiliation: Meta, Menlo Park, CA, USA.}
\endgroup

\begin{abstract}
    Regional climate risk assessment is critical for applications such as infrastructure design, disaster forecasting, and insurance resource allocation. However, estimating regional (i.e., high-spatial-resolution) risk with global climate models (GCMs) remains computationally prohibitive, which has driven the development of downscaling methods for coarse GCM outputs. Downscaling is particularly important for rare events, since quantifying their extreme properties requires high spatial resolution and very long GCM simulations. These methods non-intrusively increase GCM resolution while correcting statistical biases from unresolved fine-scale processes, thereby improving the accuracy of extreme event statistics with long return periods. A key challenge is preserving dynamical consistency, as freely evolving GCM trajectories are not expected to track the observational dataset used for training the correction operator. This is critical for causal extreme event analyses, where storyline-based risk assessment, i.e., extreme event catalogs, is necessary for effective planning. We address this challenge by introducing Dynamically and Statistically Consistent downscaling (DySCo), a non-intrusive framework that obtains high-resolution climate projections consistent with coarse GCM dynamics. DySCo relies on a data-driven reformulation of nudging to create dynamically paired training trajectories without intrusive GCM modifications. Using these paired trajectories, we train a dynamically and statistically consistent, two-stage operator. We evaluate the method by downscaling the Community Earth System Model v2 Large Ensemble (LENS2) in time and space towards historical reanalysis. Results demonstrate that DySCo operators achieve superior dynamical consistency with the coarse GCM trajectories, essentially applying minimal and causal correction to the GCM, while maintaining top statistical performance comparable to state-of-the-art unsupervised models.
\end{abstract}

\vspace{1em}
\noindent \textbf{Keywords:} Climate downscaling, Climate modeling, Nudging, Dynamical systems

\section*{Significance Statement}
Climate-sensitive applications often require localized estimates of risk, yet generating these using global climate models remains computationally prohibitive. While current statistical downscaling methods can efficiently increase data resolution, they frequently disrupt the underlying physical consistency of the trajectories. This limits their reliability for causal analysis of extreme events. We introduce DySCo, a downscaling framework that overcomes this limitation. By utilizing a data-driven trajectory-pairing technique, DySCo produces high-resolution projections that faithfully preserve the dynamical storylines of the coarse GCM outputs. This non-intrusive approach drastically reduces computational barriers while delivering the dynamically consistent, localized insights essential for robust infrastructure design, disaster forecasting, and regional planning.

\section{Introduction}

On decadal timescales, proactive economic planning, informed policy-making, and resource allocation require actionable climate risk assessments. While Global Climate Models (GCMs) \cite{Eyring2016} provide robust projections for large-scale trends, their computational cost prohibits them from operating at the finer regional scales necessary for decision-making. Planning domains, including infrastructure design \cite{MODERNIZING2024}, energy system planning \cite{Qiu2024}, flood forecasting \cite{Nearing2024}, and financial services (e.g., insurance allocation) \cite{Mills2005}, all rely on these regional projections. This challenge is further exacerbated by compound extreme events, which involve complex spatiotemporal correlations that GCMs cannot resolve \cite{Bevacqua2023, Gettelman2025, Zscheischler2018}. Consequently, bridging this scale gap remains a substantial scientific and computational challenge, with significant implications for private and federal decision-making.

The fundamental downscaling challenge lies in characterizing fine-scale physical processes driven by coarse-scale GCMs while accounting for inherent model biases and the probabilistic nature of the super-resolution problem. This task is especially difficult because the climate is a chaotic dynamical system and temporally aligned datasets are not readily available for supervised learning. Physics-based approaches employ Regional Climate Models (RCMs) \cite{Giorgi2019, Skamarock2019ADO} to resolve fine-scale climate processes regionally, providing valuable information at impact-relevant scales. However, the high computational cost of generating large ensembles with RCMs limits their risk-assessment capabilities, particularly for $O(100)$-year extreme events \cite{Goldenson2023, Giorgi2019}. Traditional statistical downscaling methods based on quantile mapping and spatial interpolation \cite{Hayhoe2024, Wood2002} are a computationally efficient alternative, but often fail to capture spatiotemporal correlations required for adequate extreme risk assessment \cite{chandel2024state}. Recently, machine learning (ML) has emerged as a powerful way to learn complex non-linear mappings between spatial scales, matching the skill of physics-based approaches at a small fraction of the cost \cite{Mardani2025NVIDIA, Lockwood2024, Lopez_Gomez_2025, Rampal2024}. However, many of these methods rely on high-resolution training data being temporally aligned with the coarse model data. This typically requires access to high-fidelity RCM data when training emulators \cite{Lopez_Gomez_2025}, or the ability to generate new GCM simulations nudged to align with observational data \cite{BarthelSorensen2024}, which can limit their broad application.

To bypass this paired-data requirement, a state-of-the-art unsupervised generative AI framework, GenFocal \cite{wan2026regionalclimateriskassessment}, was introduced. It decomposes the problem into a debiasing and a super-resolution stage, and is able to sample physically-coherent downscaled trajectories, conditioned on coarse GCM output. While this approach captures complex spatiotemporal features, moving away from the supervised learning paradigm sacrifices direct dynamical coherence with the GCM driver. As a result, while the learned mapping can generate physical climate trajectories that aggregate to the correct target statistics, it does not explicitly enforce a strict cause-and-effect relationship between the large-scale GCM forcing and localized weather. This is a fundamentally important feature to preserve for applications such as storyline-based extreme event analysis \cite{Shepherd2019, Trenberth2015}, where regional impacts must be attributed to the specific GCM driver underpinning the analysis.

In this work, we demonstrate a framework for posing climate downscaling as a supervised learning problem without the need for costly RCM or nudged GCM simulations, circumventing the practical limitations of prior methods.
The framework achieves this by learning a computationally efficient GCM emulator from coarse GCM output, then generating paired training data by nudging the emulator toward a coarsened weather reanalysis dataset. Such a process allows us to formulate a supervised learning problem from unpaired GCM outputs and reanalysis or observational targets. Machine learning models can then solve this problem in a flexible, model-agnostic, and scalable way. Furthermore, the approach ensures that generated fine-scale extremes remain explicitly and dynamically coherent with their large-scale atmospheric drivers without sacrificing computational resources.
We call the introduced framework Dynamically and Statistically Consistent (DySCo) downscaling.

To showcase the performance of our framework, we use DySCo to downscale climate projections from the CESM2 Large Ensemble (LENS2) over the Contiguous United States (CONUS) to the resolution of the European Center for Medium-Range Weather Forecasts Reanalysis v5 (ERA5) \cite{Hersbach2020}. We evaluate our method by comparing the statistics of the full 100-member downscaled ensemble against ERA5 data over a period outside the training record, and by assessing its ability to maintain dynamical coherence with the coarse GCM driver.


\section{Methods}\label{sec:methodology}

We rely on two widely used, publicly available climate datasets: the Community Earth System Model v2 Large Ensemble (LENS2) \cite{Rodgers2021} as the coarse GCM driver and the ERA5 dataset as the high-resolution reference. The former provides a comprehensive ensemble, capturing the global climate internal variability of the Earth climate model, whereas the latter provides an assimilated dataset of historical observational data at the target high resolution. From these datasets we aim to downscale four fundamental surface climatological variables: 2m temperature $T$ (K), 1000hPa specific humidity $Q$ (g/kg), 10m wind speed $W$ (m/s) and mean-sea-level pressure $P$ (Pa).

Specifically, given the $1.5^\circ$ ($\sim$ 167km at the equator), daily, low-resolution LENS2 ensemble as input, we aim to learn a non-intrusive local correction operator that debiases the input towards ERA5 and super-resolves it to the target $0.25^\circ$ ($\sim$ 28km at the equator) and 2-hourly resolution. The correction operator is localized to CONUS, covering an area approximately of size $60^\circ\times 30^\circ$ in longitude and latitude, over the summer months (June-July-August). For training, we utilize data from $1980-1999$ across both ERA5 and $4$ out of the 100 LENS2 ensemble members. The $2000-2009$ decade is used as the validation period, while the $2010-2019$ period across all 100 LENS2 members is used as the testing domain. Consequently, the model is trained on $20$ simulation years ($20\ \text{years}\times 4\ \text{members}\times 3/12\ {\rm months}$) and evaluated on $250$ simulation years ($10\ \text{years}\ \times 100\ \text{members}\times 3/12\ {\rm months}$).

\subsubsection{Problem decomposition: debiasing and super-resolution}

Downscaling coarse GCM outputs can be decomposed into two distinct tasks, debiasing and super-resolution, which have unique sets of challenges \citep{wan2026regionalclimateriskassessment}. In particular, climate projection ensembles such as LENS2, by virtue of originating from a low-resolution model, are imperfect and thus exhibit systematic biases compared to observed weather patterns \cite{Maraun2017}. These biases need to be corrected in a physically consistent way to provide statistically reliable and realistic information. However, because we can only rely on historically-constrained data products to represent the true climate, we must learn to correct this bias from a single weather trajectory. 
Further, since the climate system is chaotic, debiasing methods must differentiate biases to be corrected from trajectory divergence. This may be done through statistical aggregation or by creating climate trajectories that follow the historical weather evolution.

Next, {\color{black} the increase in spatiotemporal resolution (which in the present case is 6-fold spatially and 12-fold temporally), renders the deterministic mapping problem ill-posed, since a single input can correspond to a large number of high-resolution realities.} It should be noted that the increase in resolution is critical, since extreme events of interest (e.g. tropical cyclones) occur in smaller spatiotemporal scales than those captured by the input resolution \cite{Roberts2020}.

Finally, the combination of multi-decadal timelines, large ensemble sizes, and the multitude of variables required to accurately capture Earth's climate produces datasets that routinely scale into the terabytes. Because computational intractability is a fundamental barrier in climate research, any proposed solution must remain highly efficient to be practically viable.

Given these challenges, it becomes evident why the current problem can be naturally decomposed into a debiasing and a super-resolution stage. We adopt the same decomposition strategy as in \cite{wan2026regionalclimateriskassessment, Wood2002}, where the debiasing stage occurs at the low-resolution scale, while an independent super-resolution map is learned and composed subsequently. In particular, the target dataset $z$ is first downsampled to the input's resolution to generate an intermediate dataset $y$. Then, a debiasing map, $x\to y$, is learned at low resolution and an independent super-resolution map $y\to z$ is trained to accept debiased inputs and super-resolve them. {\color{black} For the former step, in DySCo we develop and utilize a fully data-driven variant of nudging \cite{BarthelSorensen2024} that does not require the input GCM to be resolved with external forcing, which will prove critical to maintaining dynamical coherence with the LENS2 driver, in a computationally tractable way. Therefore}, the debiasing problem can be solved at a low-cost low-resolution domain, and the super-resolution component is only responsible for solving the probabilistic resolution problem by assuming perfectly debiased input during training.

\subsubsection{The nudging-based supervised learning problem}

\begin{figure}[htb]
    \centering
    \includegraphics[width=0.6\linewidth]{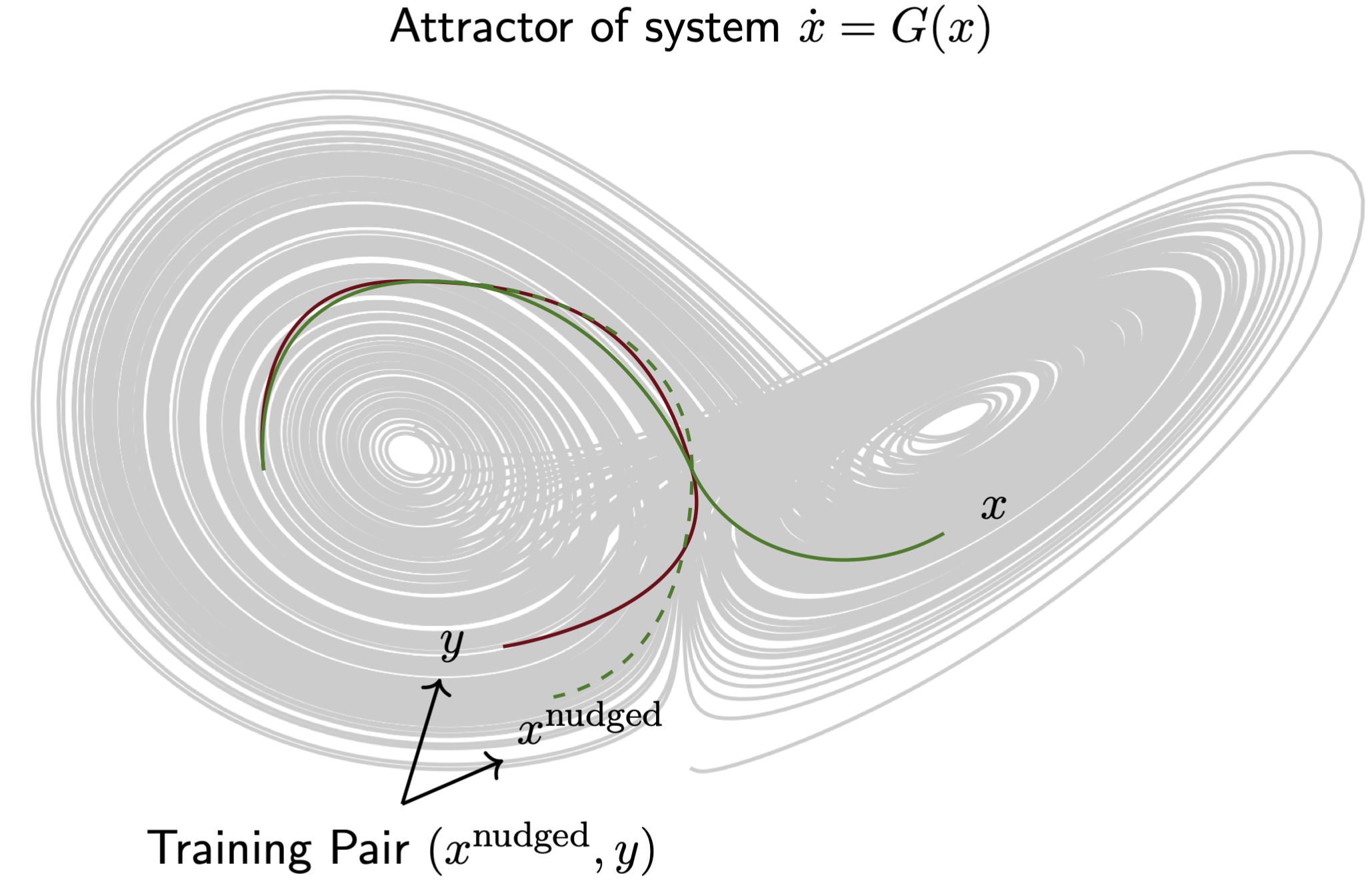}
    \caption{Illustration of the generation of paired training trajectories through nudging. Based on the original dynamical system $G$ and a reference trajectory $y$, a new dynamical system $G^{\rm nudged}$ (\ref{eq:nudging}) is defined to generate trajectories $x^{\rm nudged}$ that track the evolution of $y$. These trajectories can be used as training pairs $(x^{\rm nudged}, y)$ to define a supervised learning problem for the mapping $x^{\rm nudged}\to y$. Because the fast dynamics of $x^{\rm nudged}$ evolve freely according to $G$, the learned mapping can be applied to new trajectories $\dot{x}'=G(x')$.}
    \label{fig:attractor}
\end{figure}

Let $G$ be the dynamical system governing the LENS2 $x$ trajectories, $\dot{x}=G(x)$, and $y$ be the reference reanalysis trajectory. We can then define the corresponding nudged dynamical system \cite{BarthelSorensen2024, wang2025gen2generativepredictioncorrectionframework}, $G^{\rm nudged}$ as
\begin{align}\label{eq:nudging}
    \dot{x}^{\rm nudged}=G^{\rm nudged}(x^{\rm nudged})=G(x^{\rm nudged})-\frac{1}{\tau}(x^{\rm nudged}-y)
\end{align}
where $\tau\in\mathbb{R}$ is a user-defined relaxation timescale. At an intuitive level, the nudging $\tau$-term in Equation (\ref{eq:nudging}) has negligible effect when the trajectory $x^{\rm nudged}$ is evolving close to the reference trajectory $y$, and is activated only when these two trajectories deviate, at which times it pushes the evolving trajectory back towards $y$. These deviations occur even with near-identical initial conditions, due to the chaotic nature of the underlying systems. More precisely, it can be shown \cite{BarthelSorensen2024} that with an appropriately tuned $\tau$-term, the fast dynamics of $x^{\rm nudged}$, {\color{black} as separated by $\tau$,} are free to evolve according to the dynamical system $G$ while its slow dynamics are driven towards the trajectory $y$. Under the nudging formulation, the pair $(x^{\rm nudged}, y)$ can be used to define a supervised learning problem for a mapping from $x$ to $y$, which can be readily applied to new trajectories driven by $G$. The sketch in Figure \ref{fig:attractor} illustrates this procedure.

A critical algorithmic choice is the user-defined parameter $\tau$, which balances the faithfulness of the solution $x^{\rm nudged}$ in (\ref{eq:nudging}) to the original dynamics $G$ (as $\tau\to\infty$) against the suppression of chaotic divergences (as $\tau\to 0$). In practical terms, if $\tau$ is too large, then the trajectories will diverge, meaning that the pair $(x^{\rm nudged}, y)$ will no longer be dynamically coherent, which is fundamental to the supervised problem. If $\tau$ is not large enough, then the generated trajectory will deviate too far from the original dynamics $G$, meaning that a map trained on pairs  $(x^{\rm nudged},y)$ will not generalize to $x$ inputs. {\color{black} Optimal values of $\tau$ are problem dependent, constrained by the magnitude of chaotic divergences between $x$ and $y$, versus structural biases introduced by altering the underlying system to $G^{\rm nudged}$ that need to be overcome by the learning mapping.}

\subsubsection{Non-intrusive nudging via data-driven emulation}

The principal difficulty in using (\ref{eq:nudging}) to define a supervised problem is that in order to generate $x^{\rm nudged}$ trajectories, the original $G$ system has to be re-implemented with the nudging term present, and then evolved in time. This can be difficult when $G$ represents a GCM, and even more difficult when the goal is to downscale an ensemble of climate models developed by different organizations using diverse computational frameworks. Here, our input trajectories $x$ are given by the LENS2 ensemble, which would require modifying the CESM2 codebase and running inference with it. This would be a labor and computationally intensive task, especially considering the fact that $\tau$ must be experimentally identified. To circumvent this issue, we introduce a new methodology for generating the paired trajectory (\ref{eq:nudging}) that is fully-data driven and $G$-agnostic. Given a dataset $x$, this method relies on learning an efficient surrogate model $G_{\rm surr}$ that captures the fast-dynamics of $x$. The paired trajectories are then generated by nudging the surrogate model, instead of the original system:

\begin{align}\label{eq:emulator_nudging}
G^{\rm nudged}_{\rm surr}(x^{\rm nudged}_{\rm surr}) = G_{\rm surr}(x^{\rm nudged}_{\rm surr})-\frac{1}{\tau}(x^{\rm nudged}_{\rm surr}-y)
\end{align}

Crucially, the method capitalizes on the fact that the requirements from the surrogate $G_{\rm surr}$ are less strict than in a traditional surrogate modeling paradigm, since in the present case, $G_{\rm surr}$ need only capture the fast dynamics of $x$ to the extent to which a subsequent map can generalize these features from $x^{\rm nudged}_{\rm surr}$ to $x$. There is an interplay between the quality of $G_{\rm surr}$ and of the subsequent map that will be trained on $(x^{\rm nudged}_{\rm surr},y)$, since a highly performant map allows for a lower-quality surrogate model. This additional methodological dimension decomposes the problem into the complexity of $G$ (e.g. more resources towards a high-quality $G_{\rm surr}$) and the complexity arising from the differences between the attractors of $x$ and $y$ (e.g. more resources towards a high-quality map $x\to y$).

In this connection, a promising candidate for $G_{\rm surr}$ is the multivariate Gaussian emulator introduced in the GEN2 framework \cite{wang2025gen2generativepredictioncorrectionframework}, which has been validated for GCM emulation tasks. In particular, given a training LENS2 time series ensemble $x(t)\in \mathbb{R}^{N_{\rm ens}\times N_{\rm var}\times N_{\rm lat}\times N_{\rm lon}}$, its day-of-year climatological mean $\bar{x}_{v,i,j}(t)$, aggregated over time and the ensemble, is removed to extract the fluctuation fields $\widetilde{x}(t)$. Then, for $\sigma_{v}$ the globally averaged standard deviation for each variable $v$, the first $M\in\mathbb{N}$, Principal Component Analysis (PCA) modes of $\widetilde{x}(t)/\sigma$, $\phi^m_{i,j}$, are computed. The emulated trajectory $x_{\rm surr}$ is constructed as
\begin{align}\label{eq:emulator}
    x_{\rm surr}(t)=\bar{x}(t)+\sigma\sum_{m}a^m(t)\phi^m
\end{align}
where $a^m(t)$ is the time series of the PCA coefficients corresponding to the PCA modes $\phi^m$. The PCA coefficients $a^m(t)$ are then modeled as Gaussian stochastic processes conditioned on the yearly globally averaged temperature $T^{\rm avg}(t)\in\mathbb{R}$ as a driver,
\begin{align}\label{eq:emulator-pca-coeffs}
    a^m(t)=\hat{\mu}^m(T^{\rm avg})+\hat{\sigma}^m(T^{\rm avg})\hat{\eta}^m(t)
\end{align}
where $T^{\rm avg}$ is constant per year of emulation, $\hat{\mu}^m$ and $\hat{\sigma}^m$ are linear models of $T^{\rm avg}$, and $\hat{\eta}^m(t)$ are modeled as zero-mean $K$-order auto-regressive Gaussian processes, whose parameters are estimated from the training LENS2 time series. At inference time, given a scalar time-series $T^{\rm avg}$, $\eta(t)=[\eta^m(t)]_m$ is emulated and then the full physical climate fields $x_{\rm surr}(t)$ are reconstructed using (\ref{eq:emulator}, \ref{eq:emulator-pca-coeffs}). 

Next, (\ref{eq:emulator}) is reformulated as a dynamical system, $\dot{\eta}=G_{\rm surr}(\eta)$, and is nudged towards the fluctuations $\eta^y$ corresponding to the target reanalysis dataset, to define the nudged system,
\begin{align}\label{eq:emulator-nudged}
    \dot{\eta}^{\rm nudged}=G_{\rm surr}(\eta) - \frac{1}{\tau}(\eta^{\rm nudged}-\eta^y).
\end{align}
where {\color{black} $\eta^y$ are obtained from the physical ERA5 low resolution fields $y$, by Equations (\ref{eq:emulator}, \ref{eq:emulator-pca-coeffs}) using $y$ instead of $x$.} $G_{\rm surr}(\eta)$ is approximated with an Euler scheme and computed from the free-running emulator simulation (\ref{eq:emulator}). As in \cite{wang2025gen2generativepredictioncorrectionframework}, a final post-processing step is applied to (\ref{eq:emulator-nudged}) that mitigates artificial dissipation introduced through the nudging term by normalizing the nudged trajectories spatially, to the mean and standard deviation of (\ref{eq:emulator}). Notice that in contrast to the standard nudging formulation (\ref{eq:nudging}), a base forcing term $G_{\rm surr}(\eta)$ is precomputed as opposed to the online forcing $G_{\rm surr}(\eta^{\rm nudged})$ which makes the computation trivial, while maintaining sufficient fidelity for the supervised data generation step. After generating the nudged time series $\eta^{\rm nudged}(t)$, the physical fields $x^{\rm nudged}_{\rm surr}$ are again reconstructed using Equations (\ref{eq:emulator}, \ref{eq:emulator-pca-coeffs}) and can be used to train a map $x\to y$, with supervised learning methods using pairs $(x^{\rm nudged}_{\rm surr}, y)$. The user-defined parameters $\tau$, $M$ and $K$ have been experimentally set to $(6,100,15)$ which correspond to 6 hours, 100 PCA modes and 15-days of auto-regressive lag, respectively.

\subsubsection{The debiasing operator: a balance between statistical and dynamical learning}

We model the debiasing operator using rectified flow (ReFlow) \citep{liu2022flowstraightfastlearning}, following its successful application in unpaired climate debiasing tasks \citep{wan2026regionalclimateriskassessment}. To exploit the supervised signal provided by the data pairs $(x^{\rm nudged}_{\rm surr}, y)$, we reformulate ReFlow for a supervised learning context. In the standard ReFlow framework, which typically addresses the unsupervised problem of mapping $x \to y$ from unpaired data, an ordinary differential equation (ODE) $G_\theta$ is parameterized by learnable weights $\theta$ in fictitious time $s \in [0,1]$, such that
\begin{align}\label{eq:reflow-ode}
    \frac{d w(s)}{ds} &= G_\theta(w(s), s), \quad \text{with} \quad w(0)=x \quad \text{and} \quad w(1)=y.
\end{align}
The resulting debiasing map $G_{\rm deb}$ is then defined by integrating over this time domain:
\begin{align}\label{eq:reflow-deb-map}
    G_{\rm deb}(x) = x+\int_0^1 G_\theta(w(s), s) \, ds.
\end{align}
In its original formulation, ReFlow minimizes the objective $L(\theta) = \mathbb{E}_{s\sim \mathcal{U}[0,1]} \|y - x - G_\theta(sy + (1-s)x, s)\|^2$, which drives the transformation $x \to y$ along a linear path, where $\mathcal{U}[0,1]$ denotes the uniform distribution over $[0,1]$. To incorporate the explicit mapping from our nudged pairs $(x^{\rm nudged}_{\rm surr}, y)$, we augment this objective by adding the direct integration error from (\ref{eq:reflow-deb-map}) as a second penalty term:
\begin{align}\label{eq:reflow-adapted-min-problem}
     L(\theta) &= \mathbb{E}_{s\sim \mathcal{U}[0,1]} \Big[ \|y - x - G_\theta(sy + (1-s)x, s)\|^2 \Big] \nonumber \\
     &\quad + \lambda \|y - G_{\rm deb}(x)\|^2,
\end{align}
where $\lambda$ is a tunable hyperparameter that balances the distributional ($\lambda \to 0$) and supervised ($\lambda \to \infty$) learning signals. The gradients through the ODE solution are backpropagated using a differentiable finite-step ODE solver.

We parameterize $G_\theta$ as a spatiotemporal reformulation of the U-Net Vision Transformer (U-ViT) network \cite{bao2023worthwordsvitbackbone}. The architecture is identical to the one used in \cite{wan2026regionalclimateriskassessment}, with hyper-parameters summarized in the Appendix. The user-defined $\lambda$ parameter is set to $10.0$, which experimentally achieves a balance between the supervised and distributional signals. Further details are provided in the Appendix.

\subsubsection{The super-resolution operator}

Having decomposed the debiasing and super-resolution problems, the super-resolution map need only map the low-resolution target reanalysis dataset, $y$, to the full-resolution reanalysis dataset $z$. Given that the map $y\to z$ is inherently probabilistic, we follow the same approach as \cite{wan2026regionalclimateriskassessment}, where a diffusion model \cite{karras2022elucidatingdesignspacediffusionbased, song2020improvedtechniquestrainingscorebased, song2021scorebasedgenerativemodelingstochastic} conditioned on the input $y$ can be used to generate the full-resolution fields $z$. In particular, exploiting the fact that a spatial interpolation map $I(y)$ provides an approximation of the mean statistics of $z$ \citep{Lopez_Gomez_2025}, we model the residual $r=z-I(y)$, using the conditional diffusion model $G_{\rm sr}(y)$ to sample $p(r|y)$ and then reconstruct the full field as $I(y)+G_{\rm sr}(y)$. Here we have chosen $I$ to be the linear interpolant. Next, the first step in diffusion modeling, is to iteratively add noise to the targets $r+\epsilon\sigma^{\rm diff}(l)$, where $\epsilon\sim{\mathcal{N}}(0,1)$ and $\sigma^{\rm diff}(l)$ is the strength of the cumulative noise added until the current step $l\in[0,1]$. A denoising map $D_\theta$ is then learned, which given the conditioning input $y$, the step $l$, and the noisy sample $r+\epsilon\sigma^{\rm diff}(l)$, learns to recover $r$ by minimizing,
\begin{align}\label{eq:diff-loss}
    L(\theta)=\mathbb{E}_{\epsilon\sim{\mathcal{N}}(0,1)}\mathbb{E}_{l\sim{\mathcal{U}}[0,1]}\|D_\theta(r+\epsilon\sigma^{\rm diff}(l),l,y)-r\|^2.
\end{align}
This map can then be used to construct a stochastic differential equation-based sampler for $p(r|y)$, by solving
\begin{align}\label{eq:diff-sampling}
    d{\chi}=-2\frac{\big(\sigma^{\rm diff}\big)'}{\sigma^{\rm diff}}\big(D_\theta(\chi,l,y)-\chi\big)dl+\sqrt{2\big(\sigma^{\rm diff}\big)'\sigma^{\rm diff}}d\omega_l
\end{align}
for each input $y$, from $l=1$ to $l=0$ and initial condition $\chi(1)\sim{\mathcal{N}}(0,\sigma^{\rm diff}(1))$. Here, $\big(\sigma^{\rm diff}\big)'$ is the derivative of the noise schedule with respect to the diffusion time $l$ and $\omega_l$ is the standard Wiener process. The same 3D-ViT architecture as in the debiasing step is used to parameterize $D_\theta$, with the hyper-parameters used summarized in the Appendix.

\subsubsection{The end-to-end operator}

In summary, during training, given the LENS2 ensemble $x$ and high-resolution ERA5 target $z$, we obtain a coarse-resolution dataset $y$ by down-sampling $z$ to the spatiotemporal resolution of $x$. Then, an emulator $G_{\rm surr}$ (\ref{eq:emulator}) is trained on $x$ and then nudged towards $y$ to define the nudged system $G_{\rm surr}^{\rm nudged}$ (\ref{eq:emulator-nudged}). The nudged emulator is used to generate trajectories $x^{\rm nudged}_{\rm surr}$ tracking the slow dynamics of $y$. The pairs $(x^{\rm nudged}_{\rm surr}, y)$ are then used to train a debiasing map $G_{\rm deb}$ (\ref{eq:reflow-deb-map}) using the adapted ReFlow formulation (\ref{eq:reflow-adapted-min-problem}). Independently, a conditional diffusion model (\ref{eq:diff-loss}) is trained to sample $z$ (\ref{eq:diff-sampling}) conditioned on $y$. During inference, new LENS2 trajectories $x'$ are first debiased using $G_{\rm deb}$ and then super-resolved using $G_{\rm sr}$, arriving at the end-to-end DySCo map,
\begin{align}\label{eq:dysco-map}
     z_\text{DySCo}(x) = I(G_{\rm deb}(x))+G_{\rm sr}(G_{\rm deb}(x)).
\end{align}

\section{Results}

To validate our framework, we conduct an evaluation against ERA5, across the 2010–2019 testing decade using the fully unseen 100-member LENS2 ensemble, in three distinct stages. First, we demonstrate the primary advantage of our supervised correction approach: the preservation of strict physical alignment between the coarse atmospheric forcing (LENS2) and the high-resolution outputs. Then, we test the baseline distributional fidelity of our supervised approach, comparing it to the state-of-the-art GenFocal model {\color{black} and the widely used Bias Correction and Spatial Disaggregation (BCSD) \cite{OrtizBobea2021, Rode2021, Thrasher2022, Wood2002} statistical downscaling method. BCSD uses the climatology of the ERA5 target from the 1980-1999 training domain to renormalize the LENS2 ensemble output towards the ERA5 statistics (see the Appendix).} Finally, we test the method's capacity to reconstruct complex multivariate spatiotemporal extremes by analyzing North Atlantic tropical cyclone landfall distributions.

For notational convenience, each ensemble dataset $x_{e,v,t,i,j}$ is represented as a 5-tensor 
with dimensions $N_{\text{ens}} \times N_{\rm var} \times N_t \times N_{\text{lat}} \times N_{\text{lon}}$, which correspond to the ensemble member, climatological variable, time, latitude and longitude, respectively. The target reanalysis dataset, $y_{v,t,i,j}$, which consists of a single trajectory, is a 4-tensor without the ensemble dimension, with dimensions $N_{\rm var} \times N_t \times N_{\text{lat}} \times N_{\text{lon}}$.

\subsubsection{Dynamical coherence of corrected trajectories}

\begin{figure}[htb]
    \centering
    \includegraphics[width=0.6\linewidth]{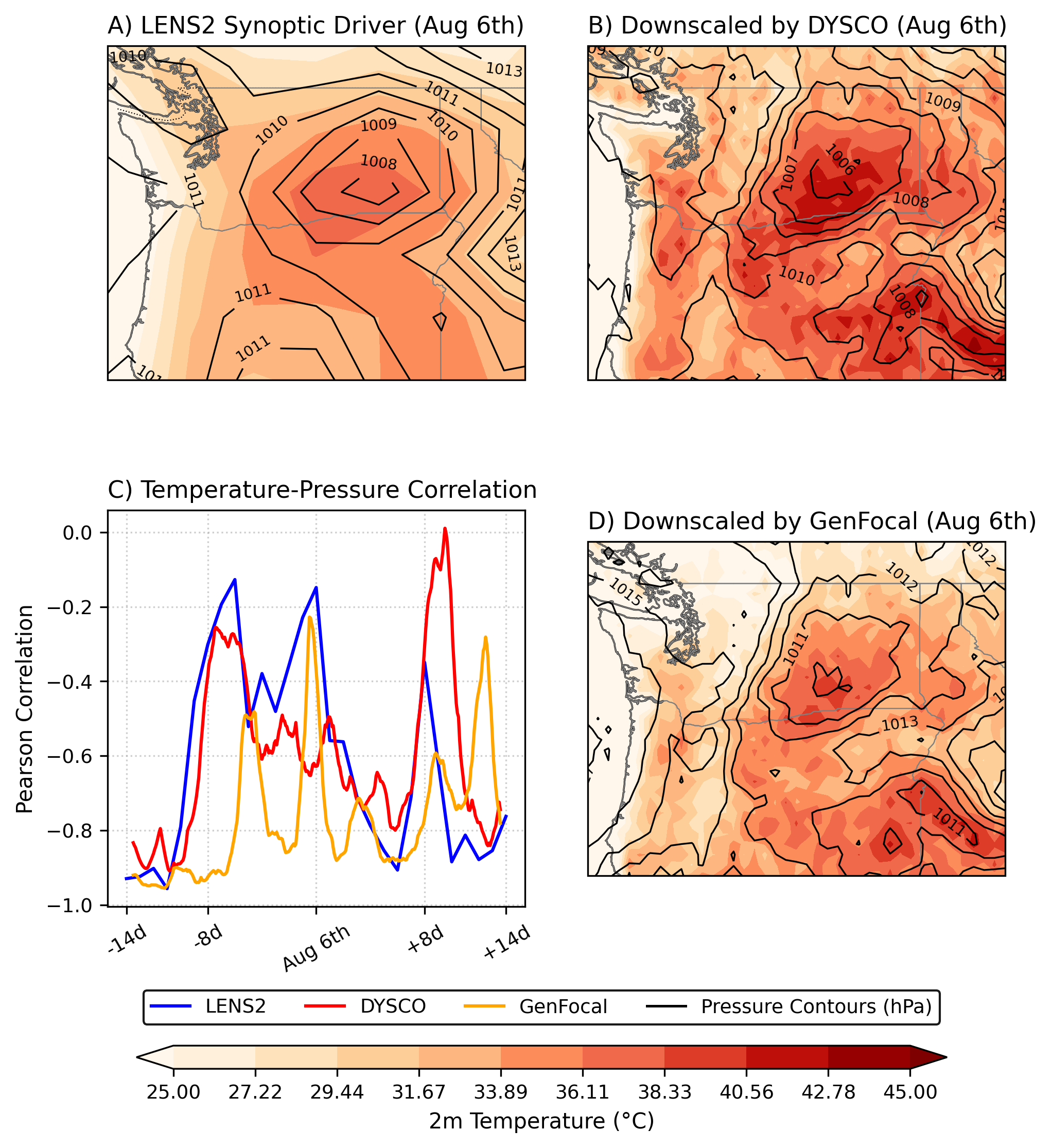}
    \caption{Case study of dynamical coherence. (A) the synoptic footprint of a LENS2 extreme temperature event in the Pacific Northwest is shown (\texttt{LE2-1251.001} member, 2016 August 6th). The contours indicate mean-sea-level pressure, revealing a surface thermal low like event ($\sim$1008hPa) co-located with extreme surface temperatures. (B) The corresponding output from DySCo, illustrating the high-resolution topographic modulation of the phenomenon, preserves the $T\leftrightarrow P$ relationship  while successfully debiasing the field toward the ERA5 distribution. (C) Temperature-pressure spatial correlation time series around the August 6th event, computed using a 1-day rolling average. (D) The corresponding output from GenFocal, which while able to sample a physical trajectory, it does not correspond to the LENS2 extreme event.}
    \label{fig:_203_MSL_LE2-1191.010_2010-06-02}
\end{figure}

We begin by validating the dynamical coherence between the GCM driver and downscaled fields. As a case study, Figure \ref{fig:_203_MSL_LE2-1191.010_2010-06-02} illustrates a Pacific Northwest heatwave simulated by LENS. Panel (A) displays the synoptic footprint, demonstrating a nonlinear inverse relationship between temperature and pressure. This is characterized by the co-occurrence of high temperatures and low pressures (~1008 hPa), indicative of a surface thermal low event \cite{Brewer2012, Bumbaco2013, Rowson1992, esd-13-1689-2022}.

Panel (B) shows that DySCo preserves this underlying $T\leftrightarrow P$ relationship, successfully resolving the fine-scale topographic modulation of the phenomenon while remaining strictly aligned with the LENS2 GCM driver. In contrast, while GenFocal generates physically realistic downscaled fields (Panel D), its dynamical evolution diverges significantly from that of the GCM driver (Panel C). This case study highlights DySCo's distinct advantage: it achieves high distributional fidelity while robustly preserving the physical signal from the driver.

Towards a quantitative comparison across models, we proceed with the definition of the Anomaly Correlation Coefficient (ACC). Because the input and output have different temporal representations (daily-averaged vs. instantaneous), the output is first aggregated to a daily mean. Furthermore, to remove the influence of the seasonal cycle, we compute correlations using climatological anomalies, which are obtained by subtracting the day-of-year climatological mean $\bar{x}_{e,v,i,j}$ from the datasets. For each variable $v$ and location $(i,j)$, we define ACC as
\begin{align}\label{eq:ACC}
    {\rm ACC}_{v,i,j}=\frac{1}{N_{\rm ens}}\sum_e\Big|\rho(\widetilde{x}_{e,v,:,i,j},\widetilde{x^{\rm out}}_{e,v,:,i,j})\Big|
\end{align}
where $\widetilde{x}$ is the input dataset climatological anomaly, $\widetilde{x^{\rm out}}$ is the debiased and super-resolved output anomaly for the low-resolution input $x$ and $\rho$ is the correlation coefficient. Given that the spatial resolution of $\widetilde{x^{\rm out}}$ is higher than $x$, locations $(i,j)$ of the latter are sampled by bilinear interpolation. We also evaluate the intermediate debiased output, $G_{\rm deb}(x)$, in which case the spatiotemporal resolutions of $\widetilde{x}$ and $\widetilde{G_{\rm deb}(x)}$ are identical and thus (\ref{eq:ACC}) can be directly applied.

\begin{figure*}[htb]
    \centering
    \includegraphics[width=0.9\linewidth]{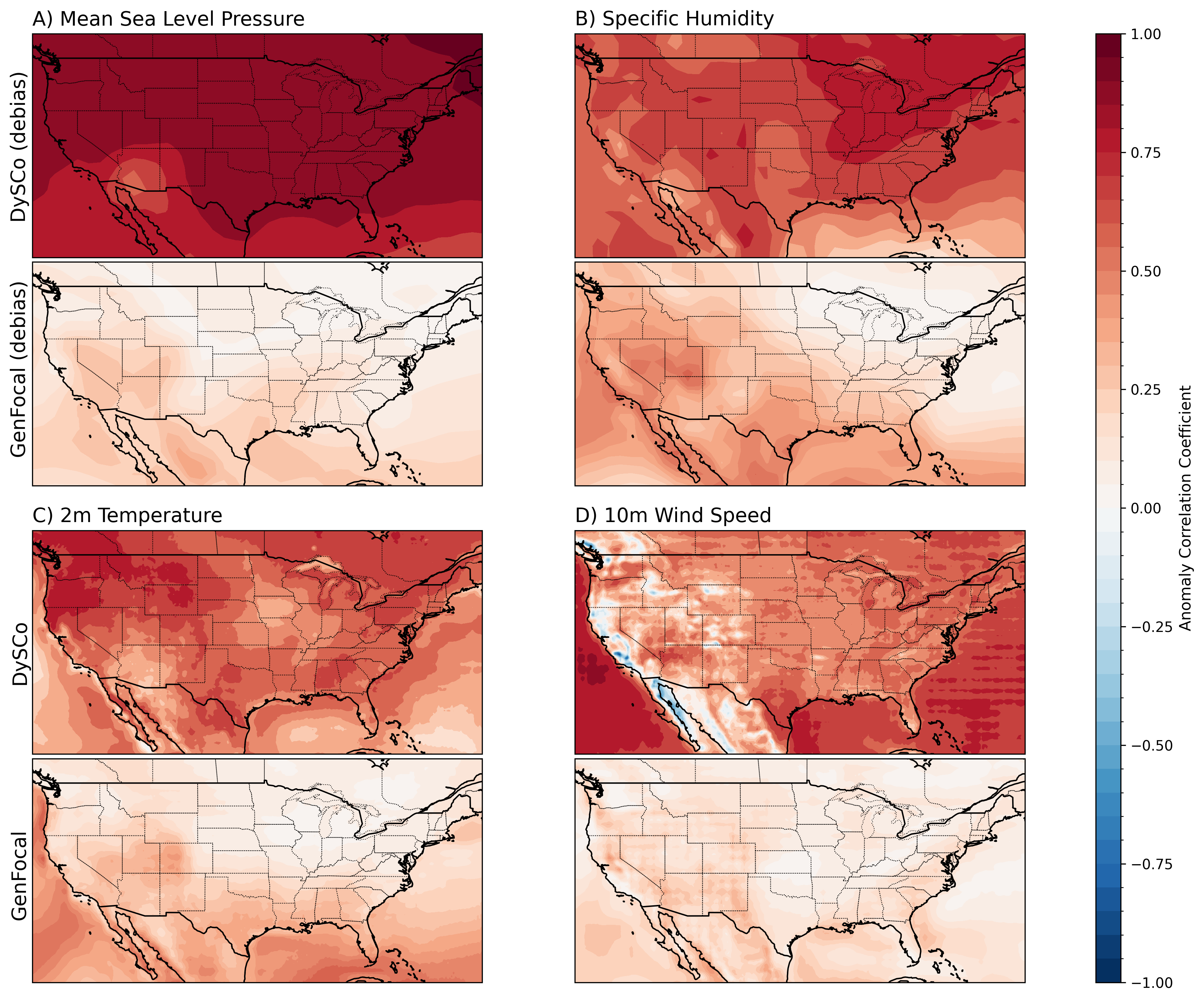}
    \caption{Anomaly correlation coefficient (\ref{eq:ACC}) between LENS2 and corrected output, for DySCo and GenFocal. The top four plots for pressure and humidity depict the correlation between LENS2 and the output of the debiasing stage of both models, $G_{\rm deb}(x)$. The bottom four plots for temperature and wind speed depict the correlation with the fully downscaled output.}
    \label{fig:_206}
\end{figure*}

Figure \ref{fig:_206} quantifies dynamical coherence in terms of the ACC (\ref{eq:ACC}) for our model and GenFocal, evaluated across the LENS2 ensemble, both for the intermediate debiased fields $G(x_{\rm deb})$, but also for the final downscaled output, in the top and bottom plots, respectively. Looking at the pressure and humidity plots, it is immediately evident that the observation in Figure \ref{fig:_203_MSL_LE2-1191.010_2010-06-02} is consistent across the ensemble, 
with DySCo showing significantly higher ACC ($\sim 0.73$) in comparison to GenFocal ($\sim 0.15$). Looking at temperature and wind speed, where we are showing the correlation between the input and the final downscaled output, this dynamical coherence is preserved by the super-resolution map, ensuring end-to-end coupling with the LENS2 driver.

\begin{figure}[htb]
    \centering
    \includegraphics[width=0.6\linewidth]{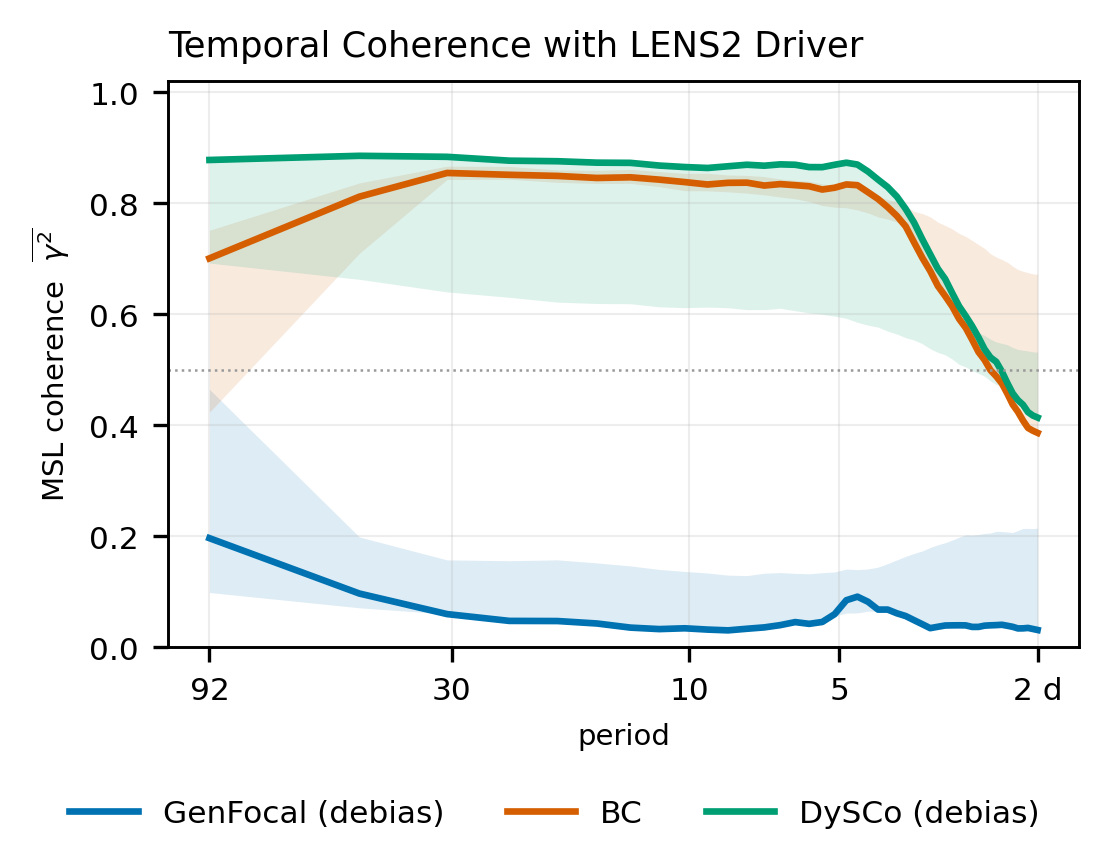}
    \caption{MSL temporal coherence between input LENS2 driver and debiased fields. Shaded region represents the 5$^{\rm th}$ and 95$^{\rm th}$ percentile range.}
    \label{fig:_430}
\end{figure}

To complement Figure \ref{fig:_206}, we evaluate temporal coherence \cite{Carter1987, vonStorch2000} as a per-frequency measure of consistency with the LENS2 driver. For each ensemble member and location, we compute coherence $\gamma^2$ between LENS2 and the output of the debiasing stage of the methods considered, aggregating these into a frequency-dependent mean coherence, $\overline{\gamma^2}$. For each frequency, $\overline{\gamma^2}$ takes values in $[0,1]$ and is a measure of phase consistency between the two signals. Values near $0$ indicate no correlation while values near $1$ indicate perfect phase synchronization. In Figure \ref{fig:_430}, MSL $\overline{\gamma^2}$ is reported over the test period for the debiasing stages of DySCo, GenFocal and BCSD. Consistently with Figure \ref{fig:_206}, DySCo retains superior coherence with the LENS2 driver than unsupervised learning methods like GenFocal. Further, BCSD does not achieve an identically $\overline{\gamma^2}=1.0$ score by construction, since even though the bias correction step is a linear transformation at each fixed time and location, the daily climatology differs non-linearly across time. It should be noted, that even though DySCo achieves higher MSL coherence than BCSD, across variables the two methods achieve comparable performance, consistently far above GenFocal. More details on the formulation of $\overline{\gamma^2}$ are provided in the Appendix.


\subsubsection{Statistical reliability of downscaled data}\label{sec:results_marginal}

To evaluate the performance of the proposed method, we proceed with a spatially marginal evaluation of the generated spatiotemporal ensemble. In particular, we consider the distribution over each individual spatial location independently, and compare it against the corresponding distribution of the reference reanalysis trajectory. The distributions aggregate across the ensemble and time axes, and are evaluated in terms of Mean Absolute Bias (MAB) (\ref{eq:MAB}), Mean Wasserstein Distance (MWD) (\ref{eq:MWD}) and Mean Percentile Error (MPE) (\ref{eq:MPE}).

MAB is defined as the time-averaged difference between the ensemble-averaged input and target, quantifies systematic deviations of the climatological variables, from the reanalysis climate. Next, MWD is defined as the mean Wasserstein distance between the input and target distributions, identifying misalignment in the bulk of the marginal probability mass. MPE, the mean error in the $p^{\rm th}$ percentile complements MAB and MWD by being able to target the tails of the underlying marginal distributions. Precise definitions of the above metrics can be found in the Appendix.

\begin{figure*}[htb]
    \centering
     \includegraphics[width=0.9\linewidth]{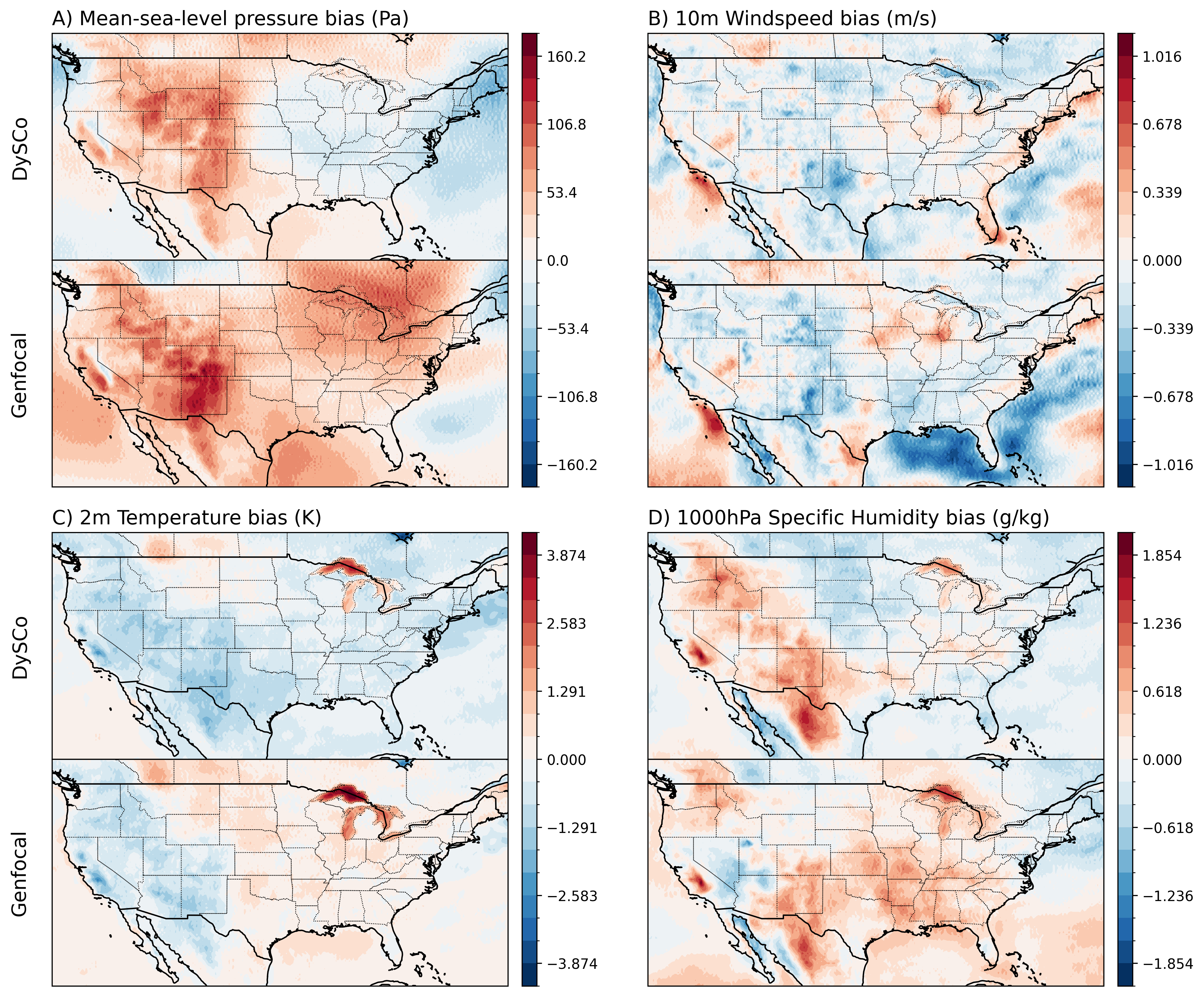}
    \caption{Bias over the 2010-2019 testing domain of the four downscaled climatological variables, for DySCo and the unsupervised GenFocal baseline.}
    \label{fig:_212}
\end{figure*}

For the testing time domain and across the LENS2 ensemble, these 3 metrics are reported in Table \ref{tab:metrics_base_variables}. {\color{black} Other than the 4 downscaled variables, we have also included the Heat Index (HI) \cite{noaa_heatindex} as a nonlinear diagnosed variable in order to evaluate the ability of the models to capture inter-variable relationships.} Recall that between our proposed method and the GenFocal baseline, the super-resolution operator is identical while the debiasing operator is 2 orders of magnitude smaller for our case (15M vs 2.6B parameters) which can be attributed to the fact that DySCo's debiasing operator is regional and the debiased fields are the same as the super-resolution inputs. In comparison to the unsupervised GenFocal baseline, we observe significantly improved performance in mean-sea-level pressure across all 3 metrics, with wind speed performing analogously or better than GenFocal while 2m temperature and specific humidity achieve either analogous or worse metrics. {\color{black} While BCSD performs consistently worse than both methods with regards to pressure, it is competitive in terms of humidity and temperature and consistently better for wind speed. Considering the trivial cost of BCSD, this level of performance is justified due to the nature of the method which while able to capture marginal statistics in time horizons where the climatology remains constant, it is not able to debias more complicated spatiotemporal features.} {\color{black}As expected, GenFocal outperforms both models in terms of the diagnosed HI variable, with DySCo's performance being in between BCSD and GenFocal.} In Figure \ref{fig:_212}, we show the bias (\ref{eq:bias}) of DySCo for the downscaled climatological variables. The improved performance of sea-level-pressure bias is reflected in this non-aggregated error map, where GenFocal appears to be biased towards lower pressure values. Similarly, the improved performance in wind speed is also apparent, with a noticeable negative bias over the east coast, while for temperature our method exhibits a bias towards lower values. {\color{black} Finally, we see the ACC performance observed in Figure \ref{fig:_206} reflected in the CONUS aggregated ACC metric, with the addition of BCSD which, by design, is also able to follow the LENS2 driver.}


\begin{table}[!ht]
\centering
\begin{tabular*}{\columnwidth}{@{\extracolsep\fill}l|cccc@{\extracolsep\fill}}
\toprule
\textbf{Variable/Model } & 
\texttt{LENS2} &
DySCo &
GenFocal &
BCSD\\
\midrule
& \multicolumn{4}{c}{Mean Anomaly Correlation Coefficient $\uparrow$} \\ 
$P$ (Pa) & - & 0.91 & 0.12 & 0.82 \\
$W$ (m/s) & - & 0.69 & 0.12 & 0.63 \\
$T$ (K) & - & 0.77 & 0.17 & 0.74 \\
$Q$ (g/kg) & - & 0.79 & 0.23 & 0.75 \\
$HI$ ($^\circ$C) & - & 0.51 & 0.13 & 0.55\\
\midrule
& \multicolumn{4}{c}{Mean Absolute Bias $\downarrow$} \\ 
$P$ (Pa) & 159.26 & 30.13 & 49.92 & 63.89 \\
$W$ (m/s) & 1.76 & 0.14 & 0.15 & 0.09 \\
$T$ (K) & 2.59 & 0.68 & 0.45 & 0.91 \\
$Q$ (g/kg) & 1.13 & 0.31 & 0.32 & 0.29 \\
$HI$ ($^\circ$C) & 2.94 & 0.70 & 0.44 & 0.90 \\
\midrule
& \multicolumn{4}{c}{Mean Wasserstein Distance $\downarrow$} \\ 
$P$ (Pa) & 162.98 & 36.88 & 59.28 & 68.53 \\
$W$ (m/s) & 1.78 & 0.23 & 0.19 & 0.20 \\
$T$ (K) & 3.07 & 0.74 & 0.54 & 0.93 \\
$Q$ (g/kg) & 1.20 & 0.42 & 0.37 & 0.35 \\
$HI$ ($^\circ$C) & 3.40 & 0.82 & 0.57 & 1.03 \\
\midrule
& \multicolumn{4}{c}{Mean Percentile Error, 99$^{\text{th}}$ $\downarrow$} \\ 
$P$ (Pa) & 161.23 & 60.64 & 78.11 & 106.38 \\
$W$ (m/s) & 1.94 & 0.35 & 0.35 & 0.27 \\
$T$ (K) & 2.73 & 0.72 & 0.67 & 0.69 \\
$Q$ (g/kg) & 1.13 & 0.70 & 0.44 & 0.44 \\
$HI$ ($^\circ$C) & 4.79 & 2.14 & 1.26 & 3.61 \\
\botrule
\end{tabular*}
\caption{Marginal metrics and ACC over CONUS, for various models.}
\label{tab:metrics_base_variables}
\end{table}

\begin{figure}[!h]
    \centering
    \includegraphics[width=0.48\linewidth]{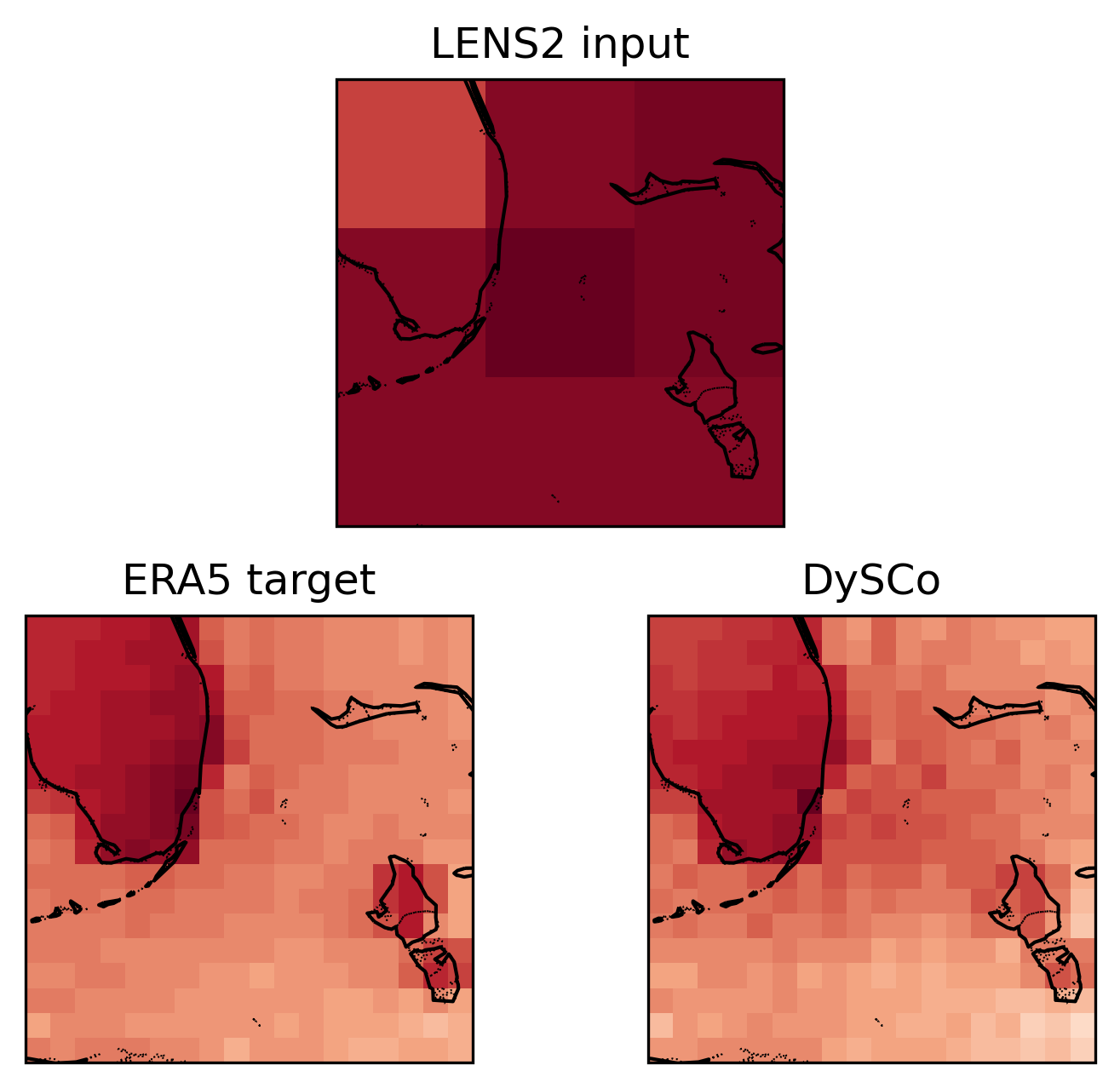}
    \caption{Spatial correlation of near-surface temperature with respect to Miami, showing the LENS2 $1.5^\circ$ input, the ERA5 $0.25^\circ$ target and the DySCo $0.25^\circ$ output.}
    \label{fig:_216_test_000_3CoRR}
\end{figure}

\begin{figure*}[htb]
    \centering
    \includegraphics[width=0.9\linewidth]{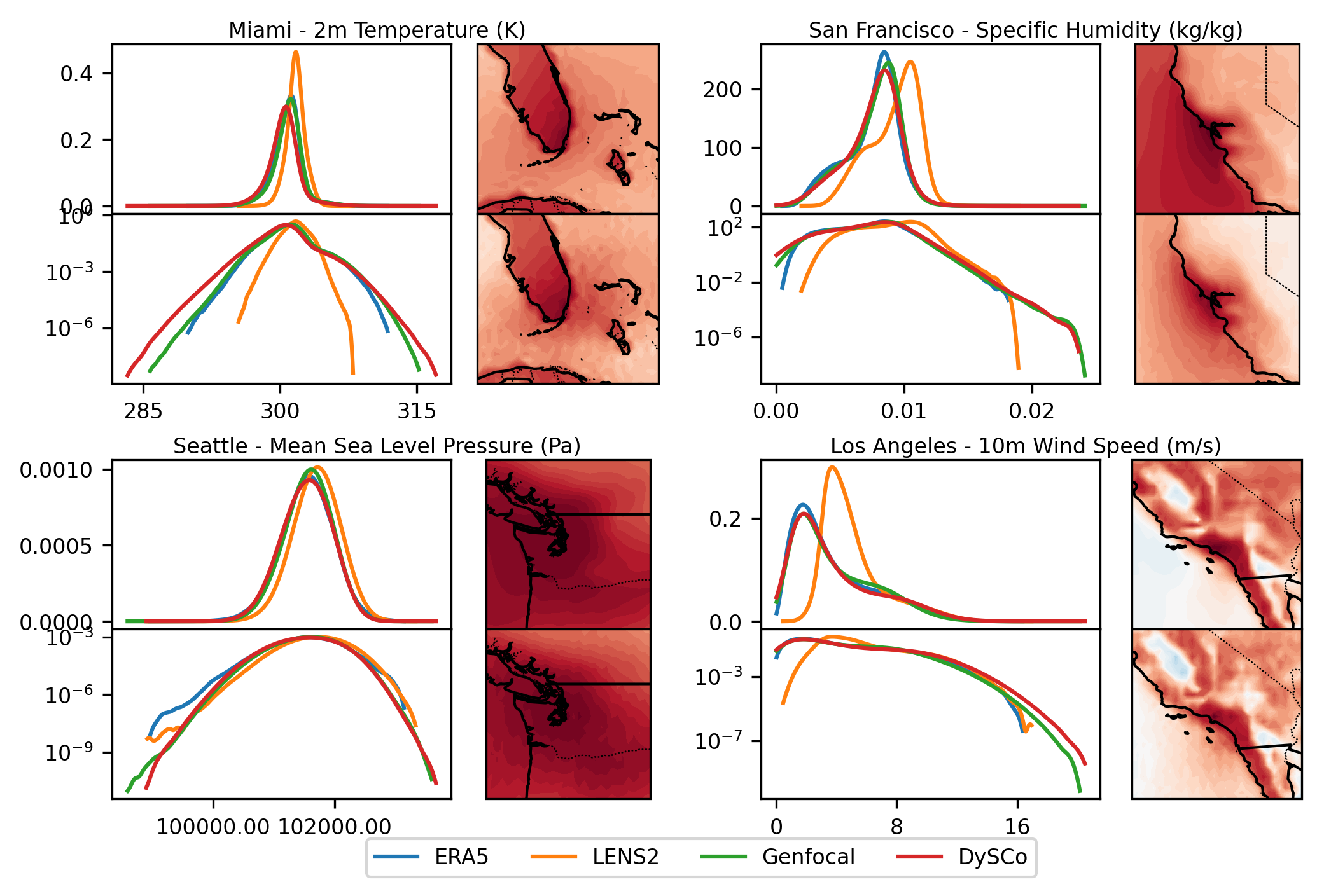}
    \includegraphics[width=0.6\linewidth]{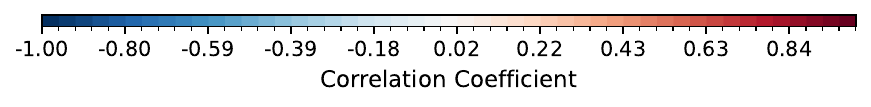}
    \caption{$4^\circ\times4^\circ$ local PDFs and correlation plots, across four different cities. For each city, top right plot is correlation of ERA5 with respect to the city location and bottom right plot is the correlation reproduced with DySCo. Top and bottom left plots are the PDFs for that variable and that region, for various models.}
    \label{fig:_216_test_010_pdfCorrMulti}
\end{figure*}

To complement the global metrics (\ref{eq:MAB}, \ref{eq:MWD}, \ref{eq:MPE}), we proceed with evaluating the local distributional performance of each climatological variable. In particular, in Figure \ref{fig:_216_test_010_pdfCorrMulti} we show the distribution of the four variables, at four cities across the US  over $4^\circ\times 4^\circ$ ($\sim (444\text{km})^2$ at the equator) neighborhoods. We also compute the corresponding local spatial correlation maps, and compare with the ERA5 target. At each correlation map, the correlation of the city's center with neighboring locations is shown. It is clear that the distributions of the downscaled variables are correctly matched to the target, in contrast to LENS2. Further, correlation structures arising from geographical features, such as the Los Angeles coastline, are correctly recovered at the target high-resolution. To further illustrate this point, Figure \ref{fig:_216_test_000_3CoRR} shows the correlation map of temperature at Miami on a smaller neighborhood, where the coastline feature that is not resolved by the $1.5^\circ$ LENS2 GCM output, is recovered by DySCo.

\subsubsection{Generated cyclone spatiotemporal trajectories}

In order to evaluate our method over complex spatiotemporal multi-variable features, we track tropical cyclone trajectories in the downscaled fields and compare them with the corresponding features in the ERA5 target and the GenFocal and BCSD baselines. We define each trajectory based on wind speed and mean-sea-level pressure, according to the criteria mimicking those in \cite{wan2026regionalclimateriskassessment}. A detailed formulation of the cyclone feature tracking algorithm is included in the Appendix.

The track probability density of these features are shown in Figure \ref{fig:_201_plot_densities_s189_into_760_275K}, where a notable low-probability region is apparent at the Gulf in the case of BCSD, whereas both DySCo and GenFocal are able to better capture the ERA5 distribution. A critical characteristic of cyclones for risk assessment and planning is whether their trajectories will make landfall and, if so, their precise points of initial impact \cite{Zhong2026landfall}. We define the landfall location of a trajectory originating over water as the first grid cell it intersects with a land fraction of at least 50\%. In Figure \ref{fig:_208_189_100K_0to99members_into_760_275K_correctStats_test_95mems__nmems51}, we plot the spatial histogram of the 250-year ($10\ \text{years}\ \times 100\ \text{members}\times 3/12\ {\rm months}$) downscaled landfall locations, from {\color{black} all ensemble models}, as well as the exact locations of the corresponding features in ERA5, for the same 2010-2019 testing period. Focusing on the Gulf, we notice {\color{black} how BCSD entirely fails to capture the landfall distribution in that region in comparison to both DySCo and GenFocal. This is of particular importance given the climatological relevance of that region and the modeling challenges it poses \cite{DurnQuesada2020}. 


\begin{figure*}[htbp]
    \centering
    \includegraphics[width=0.9\linewidth]{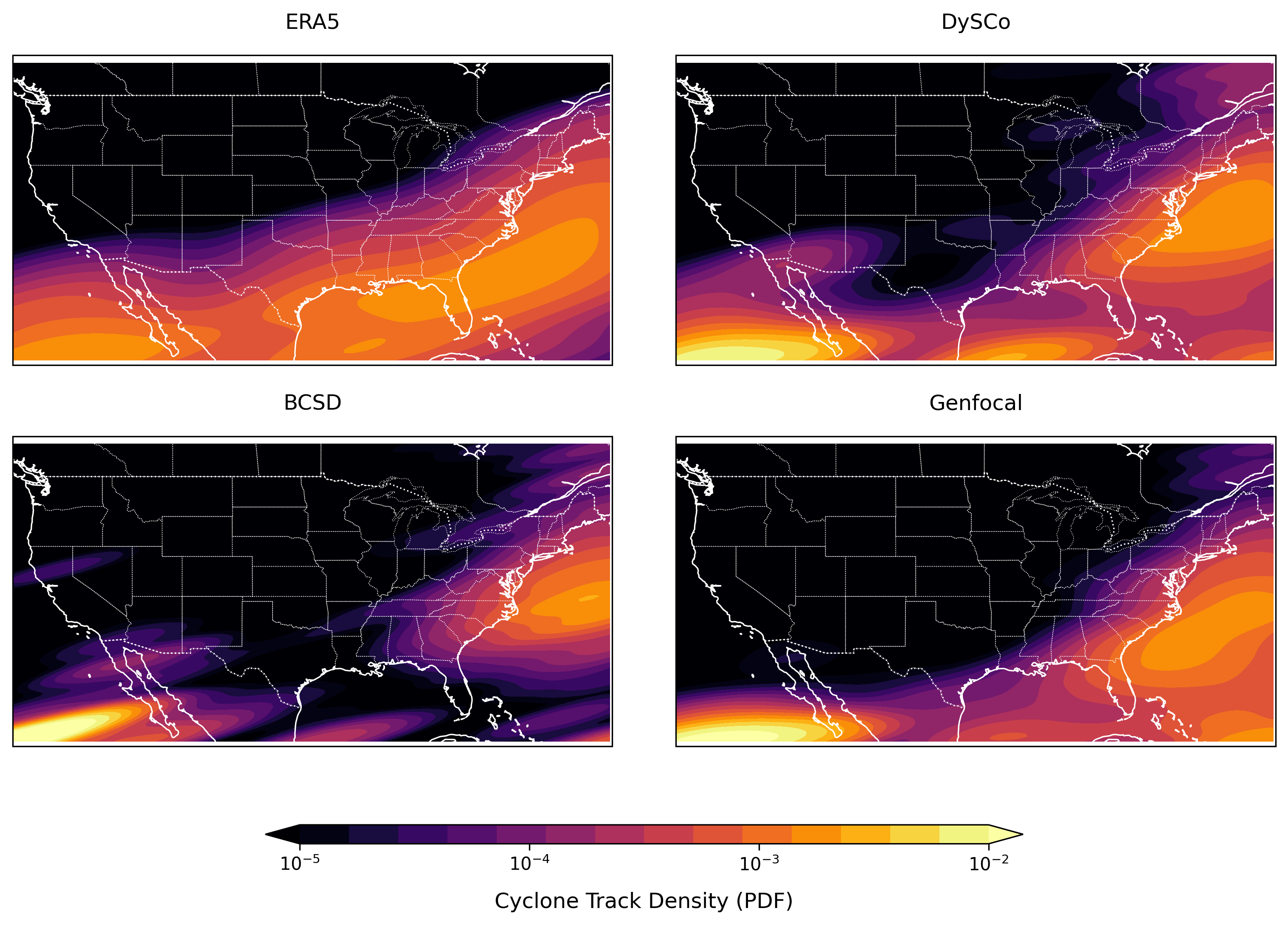}
    \caption{Probability density of cyclone trajectories {\color{black} of the ERA5 trajectory and the ensembles generated through DySCo, BCSD and GenFocal}.}
    \label{fig:_201_plot_densities_s189_into_760_275K}
\end{figure*}

\begin{figure*}[htb]
    \centering
    \includegraphics[width=0.76\linewidth]{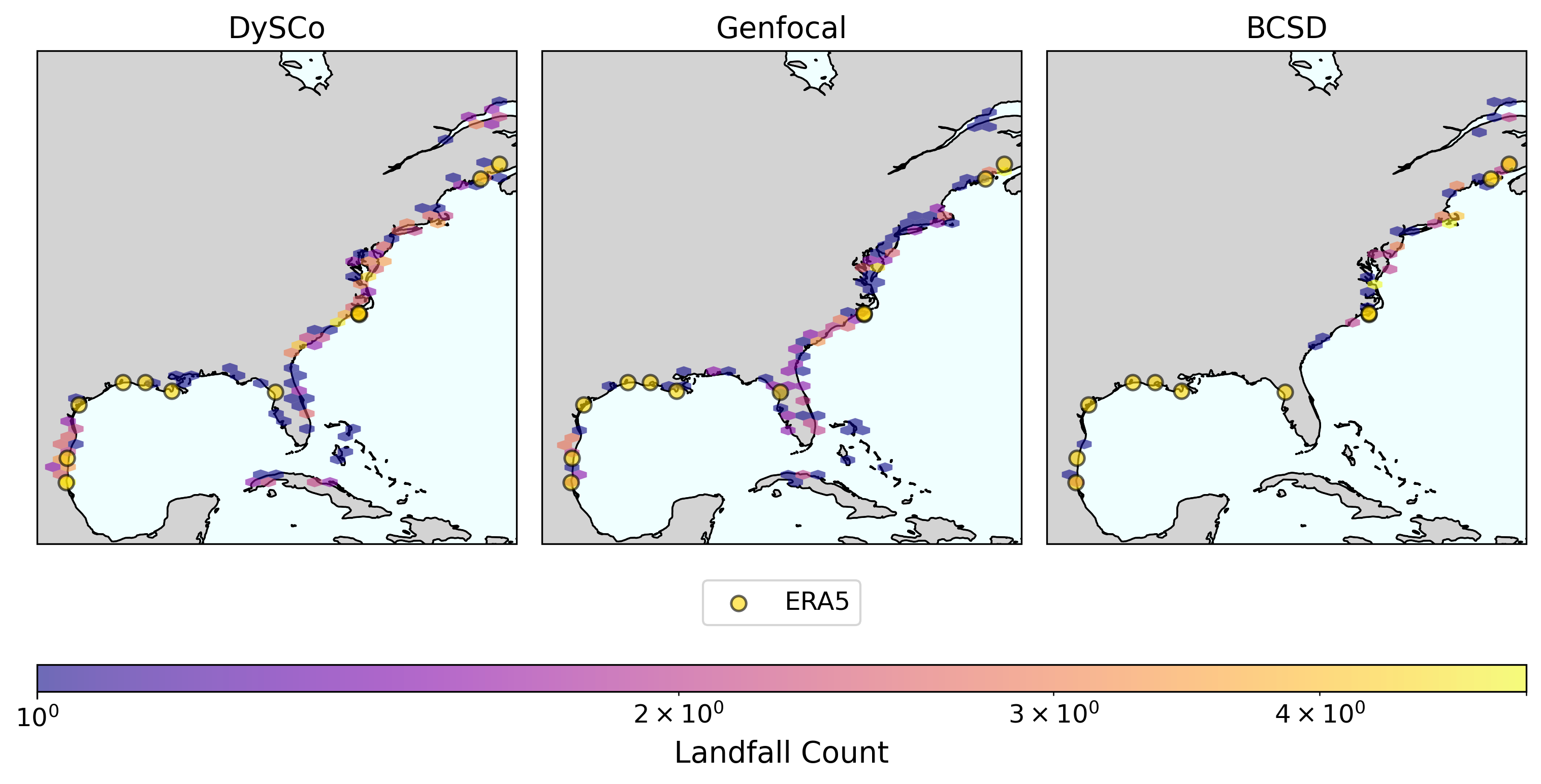}
    \caption{East-coast cyclone landfall distribution of {\color{black} DySCo, GenFocal and BCSD}, over the testing period for 100 members. The points are the ERA5 single-trajectory landfall locations.}
    \label{fig:_208_189_100K_0to99members_into_760_275K_correctStats_test_95mems__nmems51}
\end{figure*}




\section{Conclusion}


In climate downscaling problems, preserving dynamical consistency with the GCM driver is essential for causal analysis of extreme events, where effective planning relies on storyline-based risk assessments and comprehensive event catalogs driven by the GCM. In this context, we introduced DySCo, a non-intrusive, model-agnostic supervised downscaling framework that ensures both dynamical and statistical consistency.

DySCo utilizes nudging \cite{BarthelSorensen2024} and a low-cost climate emulator \cite{wang2025gen2generativepredictioncorrectionframework} to generate fully data-driven, paired trajectories for training a supervised downscaling operator. We evaluated the method on a downscaling task using the LENS2 GCM dataset as input and the ERA5 dataset as the target, achieving 6-fold spatial and 12-fold temporal increases in resolution, respectively. For this task, we implemented a supervised adaptation of GenFocal \cite{wan2026regionalclimateriskassessment}, a state-of-the-art unsupervised downscaling model, thereby merging the strengths of distribution matching and supervised learning.

In alignment with the introduced computationally efficient training data generation approach, we demonstrated that with a smaller-scale regional debiasing operator 
we are able to not only stay dynamically coherent with the LENS2 driver, but also achieve analogous performance to the larger-scale unsupervised GenFocal baseline. In fact, we have shown that the introduced approach is able to achieve better performance in the marginals of the spatiotemporal distribution for some of the climatological variables. In terms of cheaper traditional statistical downscaling approaches, we show that the BCSD method entirely misses the cyclone landfall distribution in the challenging Gulf climate, while DySCo is able to populate the coast with such landfall events.

While demonstrating this level of performance with a significantly smaller model communicates the central thesis of this work, future work can focus on applying the current approach to a model supported by industry-level resources, to match GenFocal's scale and complexity while taking advantage of the dynamically coherent signal coming from the supervised problem. In this regard, we expect a global debiasing model with more debiased climatological variables, as is the case of GenFocal, to further improve performance by providing a fuller climatological picture to the model during training and inference. Further, while the climate emulator used to define the supervised problem is a generic climate emulator that has proven to be appropriate as a nudging surrogate, developing an adapted emulator specifically designed to capture the fast dynamics of the underlying GCM should allow for higher fidelity training data. Finally, while this method has been applied to the domain of debiasing and downscaling for climate modeling, it can be directly translated to other fields such as computational fluid dynamics, or other multi-scale and multi-fidelity dynamical system-based domains.

Without requiring paired historical data, the proposed pipeline allows for the computationally tractable definition of a supervised learning problem in a fully data-driven setting. The resulting supervised DySCo operator enables the attribution of extreme events observed in the coarse GCM output to fine-scale processes as demonstrated in the case study in Figure \ref{fig:_203_MSL_LE2-1191.010_2010-06-02}. At the same time, this framework allows the user to balance computational resources based on the absolute complexity of the input and target climate models versus their dynamical dissimilarity. Ultimately, we break the supervised barrier, by introducing a method that leads to scalable and generalizable dynamically coherent ML frameworks for downscaling climate models. 

\section{Acknowledgments}
{\color{black} This project has been supported by the MIT-Google Program for Computing Innovation by the MIT Schwarzman College of Computing. We also acknowledge computational resources provided by the MIT Engaging Cluster and also the Anvil Cluster. We also thank Sheide Chammas for constructive discussions.}

\section{Competing interest}
The authors declare that they have no competing interests.

\section{Author contributions}

Stamatios Stamatelopoulos (Conceptualization, Methodology, Software, Validation, Formal Analysis, Data Curation, Writing – Original Draft Preparation, Writing – Review \& Editing, Visualization), Mengze Wang (Conceptualization, Methodology, Software), Ignacio Lopez-Gomez (Conceptualization, Methodology, Formal Analysis, Writing – Review \& Editing), Leonardo Zepeda-N\'u\~nez (Conceptualization, Methodology, Formal Analysis, Writing – Review \& Editing), Zhong Yi Wan (Conceptualization, Methodology), Robert Carver (Writing – Review \& Editing), Fei Sha (Conceptualization, Methodology, Formal Analysis, Writing – Review \& Editing, Supervision, Funding Acquisition), Themistoklis P. Sapsis (Conceptualization, Methodology, Formal Analysis, Resources, Writing – Review \& Editing, Supervision, Funding Acquisition).

\section{Data availability}
ERA5 and LENS2 data are publicly available. Public repositories for the various components of the code will be made available.

\bibliographystyle{unsrtnat}
\bibliography{reference}

\appendix

\section{Debiasing operator dependence on the supervised signal parameter}

In the adapted rectified flow minimization problem (\ref{eq:reflow-adapted-min-problem}), the $\lambda$ parameter controls the departure from the original unsupervised GenFocal formulation of the debiasing operator. In particular, $\lambda$ controls the supervised $\lambda\to\infty$, versus unsupervised $\lambda\to 0$, signal learned. Larger values of $\lambda$ weigh a direct integration term, which punishes deviations of the input from the exact reference output, a signal that is exploitable because the current setting is that of supervised learning.

In Figure \ref{fig:_219_lambda_ablation_ALL} we report the MAB (\ref{eq:MAB}), MWD (\ref{eq:MWD}) and MPE (\ref{eq:MPE}) performance of $G_{\rm deb}(x;\lambda)$ in comparison to the coarse-scale ERA5 $y$ data, during the training period for varying values of $\lambda$. We observe that performance of temperature, wind speed and humidity increases with $\lambda$, while it remains constant for pressure and decreases for the largest value of $\lambda=50$. In that way, the data hints at an optimal value of $\lambda=10$, which exploits the supervised signal to improve the performance of the first 3 variables while avoids the decrease in pressure performance.

\begin{figure}[hb]
    \centering
    \includegraphics[width=0.5\linewidth]{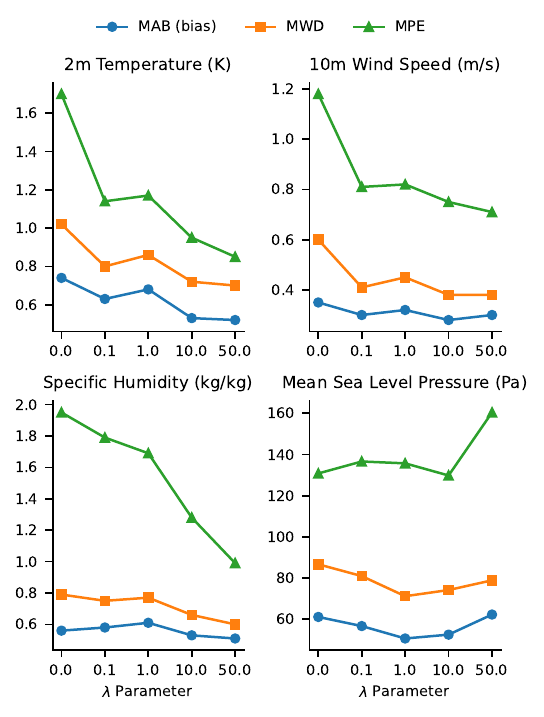}
    \caption{Dependence of the debiasing operator on the $\lambda$ parameter that control the supervised versus unsupervised signal learned. Marginal performance with respect to the metrics (\ref{eq:MAB}, \ref{eq:MWD}, \ref{eq:MPE}) is reported, for the debiased output $G_{\rm deb}(x;\lambda)$ in comparison to the coarse scale ERA5 data $y$, for $G_{\rm deb}$ trained with varying values of $\lambda$.}
    \label{fig:_219_lambda_ablation_ALL}
\end{figure}

\section{Hyperparameters}

The hyperparameters used for the debiasing operator $G_{\rm deb}$ and the super-resolution operator $G_{\rm sr}$, are summarized in Table (\ref{tab:debiasing}) and Table (\ref{tab:superres}), respectively.

\begin{table}[!t]
\centering
\begin{tabular*}{\columnwidth}{@{\extracolsep\fill}ll@{\extracolsep\fill}}
\toprule
Architecture \\ 
\midrule
Sample length & 2 days \\
Residual blocks & [3,3,3] \\
Downsampling ratios & [2,2,2] \\
Hidden channels & [64,64,64] \\
Spatial attention & [\texttt{False},\texttt{True},\texttt{True}] \\
Temporal attention & [\texttt{False},\texttt{True},\texttt{True}] \\
Trainable parameters & 15,738,132 \\
\midrule
Training \\
\midrule
Time domain & [1980,1999] \\
Ensemble domain & Members \texttt{1001\_001}, \texttt{1251\_001},\\
& \texttt{1301\_010} and \texttt{1301\_020} \\
Supervised weight & $\lambda=10$ \\
Duration & 100,000 steps \\
Dropout rate & 50\%\\
\midrule
\end{tabular*}
\caption{Debiasing model architecture and training hyperparameters.}
\label{tab:debiasing}
\end{table}

\begin{table}[!t]
\centering
\begin{tabular*}{\columnwidth}{@{\extracolsep\fill}ll@{\extracolsep\fill}}
\toprule
Architecture \\ 
\midrule
Sample length & 7 days \\
Residual blocks & [4,4,4,4] \\
Downsampling ratios & [3,2,2,2] \\
Hidden channels & [128, 256, 384, 512] \\
Spatial attention & [\texttt{False},\texttt{False},\texttt{False},\texttt{True}] \\
Temporal attention & [\texttt{False},\texttt{False},\texttt{False},\texttt{True}] \\
Sample resize & [288,144] \\
Trainable parameters & 130,791,368\\
\midrule
Training \\
\midrule
Time domain & [1980,1999] \\
Duration & 300,000 steps \\
Conditional dropout rate & 10\%\\
\midrule
\end{tabular*}
\caption{Super-resolution model architecture and training hyperparameters.}
\label{tab:superres}
\end{table}

\section{Bias Correction and Spatial Disaggregation}
{\color{black}
We adopt the multi variable spatiotemporal BCSD formulation used in \cite{wan2026regionalclimateriskassessment}, which relies on training data in the form of a climatology. In particular, given the target ERA5 dataset $z$ during the training years 1980-1999, let $y$ be the downsampled version of $z$ to the LENS2 input's $x$ resolution and $z_{\rm daily}$ be an intermediate dataset derived from $z$ by retaining the same spatial resolution, but averaging daily. Then, the first step of BCSD which is Bias Correction (BC) is to compute the debiased anomalies of $x$, as
\begin{align}
    x_{\rm BC} = \frac{x - \texttt{clim\_mean}(x)}{\texttt{clim\_std}(x)}\cdot \texttt{clim\_std}(y),
\end{align}
where \texttt{clim\_mean}$(\cdot)$ and \texttt{clim\_std}$(\cdot)$ are the climatological mean and standard deviation of the input dataset, computed for each location and day-of-year. Next, $x_{\rm BC}$ is cubically interpolated to the target spatial resolution and its mean is shifted by the climatological mean of $z_{\rm daily}$,
\begin{align}
    x_{\rm BC}'=\texttt{interp}(x_{\rm BC})+\texttt{clim\_mean}(z_{\rm daily}).
\end{align}
Finally, the last step is to disaggregate temporally, which is done by randomly sampling the historical data $z$ for a daily dataset aligned with the day-of-year entries of the $x_{\rm BC}'$ ensemble trajectories, normalizing the sample by removing its daily mean and substituting $x_{\rm BC}'$ which precisely is a daily-averaged trajectory of the climatological variables,
\begin{align}
    x_{\rm BCSD}=&z_{\rm hist\ sample} - \texttt{daily\_mean}(z_{\rm hist\ sample})\nonumber\\ +& x_{\rm BC}'.
\end{align}
}

\section{Cyclone Tracking Algorithm}

The spatiotemporal cyclone trajectories are defined based on wind speed and mean-sea-level pressure, according to the following criteria which mimic those used in \cite{wan2026regionalclimateriskassessment},
\begin{itemize}
    \item A sustained 200 Pa pressure local minimum, below 10 m elevation, in a 5.0 great circle distance (GCD).
    \item Accompanying wind speeds exceeding 10m${\rm s}^{-1}$, for a duration of at least 2 days.
    \item 8.0 GCD maximum allowable distance between consecutive trajectory nodes.
    \item 54 hour minimum trajectory duration.
    \item Maximum instantaneous speed along cyclone track of 40 knots \cite{Kossin2018, Marchok2021}
\end{itemize}
These features are identified using the open-source Tempest Extremes (TE) \cite{Ullrich2021} package with the instantaneous speed filter being applied on the TE output.

Further, given the LENS2 driver's inherent limitations in underestimating pressure depressions, we adopt a prevalent calibration approach where the output is rescaled for cyclone tracking to match ground truth cyclone statistics over a reference period in the training dataset \cite{Kochkov2024, wan2026regionalclimateriskassessment, Emanuel2021}. In particular, we select the period of [1995, 1999] and calibrate a parameter $K$ which influences the magnitude of pressure depressions via a conditional affine transformation as,
\begin{align}
     P_{\rm calibrated} = \begin{cases}
         K P+(1-K)P_{\rm amb}&,P\le P_{\rm amb}\\
        P &,{\rm else}
     \end{cases},
 \end{align}
where $P$ is the input pressure and $P_{\rm amb}$ is ambient pressure set to 1010 hPa. For each value of $K$, the cyclone count, track length and duration are computed and the relative errors in each with respect to ERA5 are summed to compute the total relative error per $K$. This procedure is repeated per model to find the corresponding optimal $1/K\in\{0.1,0.2,...,1\}$, where $1/K=1$ corresponds to no calibration. In Figure \ref{fig:_409_combined_ablation} the resulting ablation study over $K$ is shown, with identified optimal values for DySCo, BCSD and GenFocal being $1/K=0.9$, 0.9 and 1.0, respectively. 

\begin{figure}[ht]
    \centering
    \includegraphics[width=0.5\linewidth]{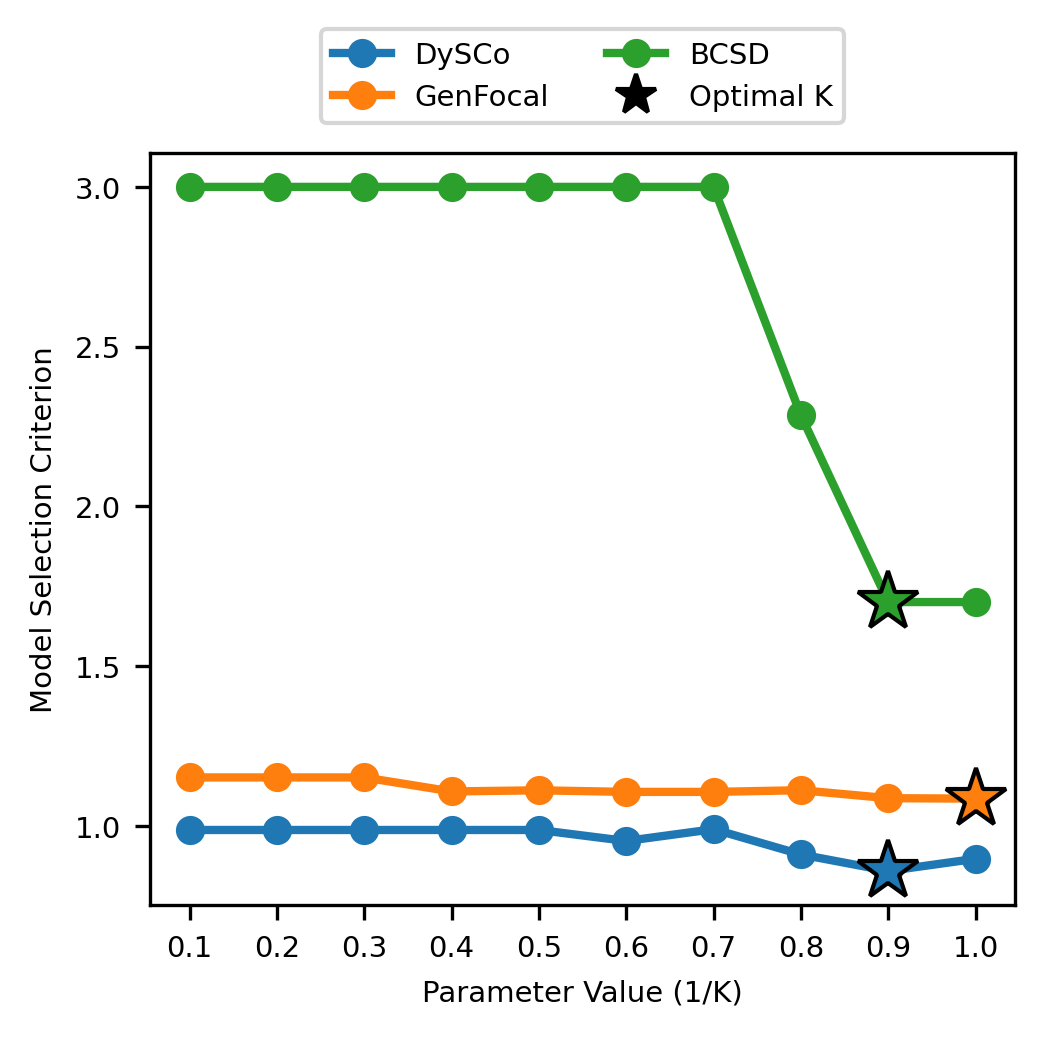}
    \caption{Per-model calibration of $K$ via an ablation study over $1/K\in\{0.1,0.2,...,1\}$.}
    \label{fig:_409_combined_ablation}
\end{figure}

\section{Marginal Metrics}

In this section we provide further details on the computation of the 3 metrics reported in Table \ref{tab:metrics_base_variables}.

Starting with MAB, for each variable $v$ and location $(i,j)$, bias is defined as the time-averaged difference between the ensemble-averaged input and target,
\begin{align}\label{eq:bias}
    {\rm Bias}_{v,i,j}=\frac{1}{N_t}\sum_{t}\Bigg(z_{v,t,i,j}-\frac{1}{N_{\rm ens}}\sum_e x_{e,v,t,i,j}\Bigg)
\end{align}
Then, for each variable $v$, mean absolute bias is the defined as the spatially aggregated ${\rm Bias}_{v,i,j}$,
\begin{align}\label{eq:MAB}
    {\rm MAB}_{v} = \frac{1}{N_{\rm lat}N_{\rm lon}}\sum_{i,j}|{\rm Bias}_{v,i,j}|.
\end{align}
MAB quantifies systematic deviations of the climatological variables, from the target reanalysis climate. To complement MAB, for each variable $v$ and at each location $(i,j)$ the Wasserstein-1 metric is computed between the corresponding distributions $x_{:,v,:,i,j}$ and $z_{v,:,i,j}$. Algorithmically, for ${\rm CDF}^b_{v,i,j}$ being the cumulative distribution function computed from either $b=x$ or $b=z$,
we define,
\begin{align}
    {\rm Wass}_{v,i,j}=\int |{\rm CDF}^x_{v,i,j}(q)-{\rm CDF}^z_{v,i,j}(q)|dq
\end{align}
where the integral is evaluated with a quadrature rule over the union of sorted quadrature points from both distributions. The mean Wasserstein distance between the 2 distributions is then defined as,
\begin{align}\label{eq:MWD}
    {\rm MWD}_{v} = \frac{1}{N_{\rm lat}N_{\rm lon}}\sum_{i,j}{\rm Wass}_{v,i,j}.
\end{align}
Finally, to complement MAB which targets systematic deviations and MWD which identifies misalignment in the bulk of the probability mass, the $p^{\rm th}$ percentile error is also evaluated, which targets the tails of the underlying distributions for large values of $p$. For ${\rm Pctl}^b_{v,i,j}$ being the percentile function computed from either $b=x$ or $b=z$, we define,
\begin{align}
    {\rm Perc}_{v,i,j}(p)=|{\rm Pctl}^x_{v,i,j}(p)-{\rm Pctl}^z_{v,i,j}(p)|
\end{align}
and as before,
\begin{align}\label{eq:MPE}
    {\rm MPE}_{v}(p)=\frac{1}{N_{\rm lat}N_{\rm lon}}\sum_{i,j}{\rm Perc}_{v,i,j}(p).
\end{align}

To produce Table \ref{tab:metrics_base_variables}, the above spatial averages have been taken over land spherically weighted by area.

\section{Temporal Coherence}

For $x$ the LENS2 ensemble and $x_{\rm deb}$ the debiased low resolution output by any of the downscaling methods, the coherence ${\rm Coh}_{e,v,f,i,j}$ at member $e$, variable $v$, frequency $f$ and location $(i,j)$ between $x$ and $x_{\rm deb}$ is defined as \cite{Carter1987, vonStorch2000},
\begin{align}\label{eq:coherence}
\gamma^2_{e,v,f,i,j}
  = \frac{\bigl|\bar{S}_{e,v,f,i,j}^{xx_{\rm deb}}\bigr|^{2}}
         {\bar{S}_{e,v,f,i,j}^{xx}\,
          \bar{S}_{e,v,f,i,j}^{x_{\rm deb}x_{\rm deb}}}
\end{align}
where $S$ is the spectral density of the time-series at $(e,v,i,j)$. These spectra are computed by the discrete Fourier transforms of the underlying signals using the standard Welch estimators \cite{WELCH19671161901}. The values of lie in $[0,1]$, with $\gamma\to 0$ indicating no phase or amplitude consistent relationship between the two signals. To summarize coherence performance, (\ref{eq:coherence}) is aggregated across $(e,i,j)$, to arrive at the metric
\begin{align}\label{eq:coherence_aggregated}
\overline{\gamma^2_{v,f}}
  = \sum_{e,i,j}w_{e,i,j}\gamma^2_{e,v,f,i,j}
\end{align}
for each variable $v$ and frequency $f$, where $w_{e,i,j}$ are the appropriate spherical weights for the $(i,j)$ aggregation, while the members $e$ are uniformly weighted. In Figure \ref{fig:_427}, we report a version of (\ref{eq:coherence_aggregated}) that is aggregated across variables.

\begin{figure}[htb]
    \centering
    \includegraphics[width=0.5\linewidth]{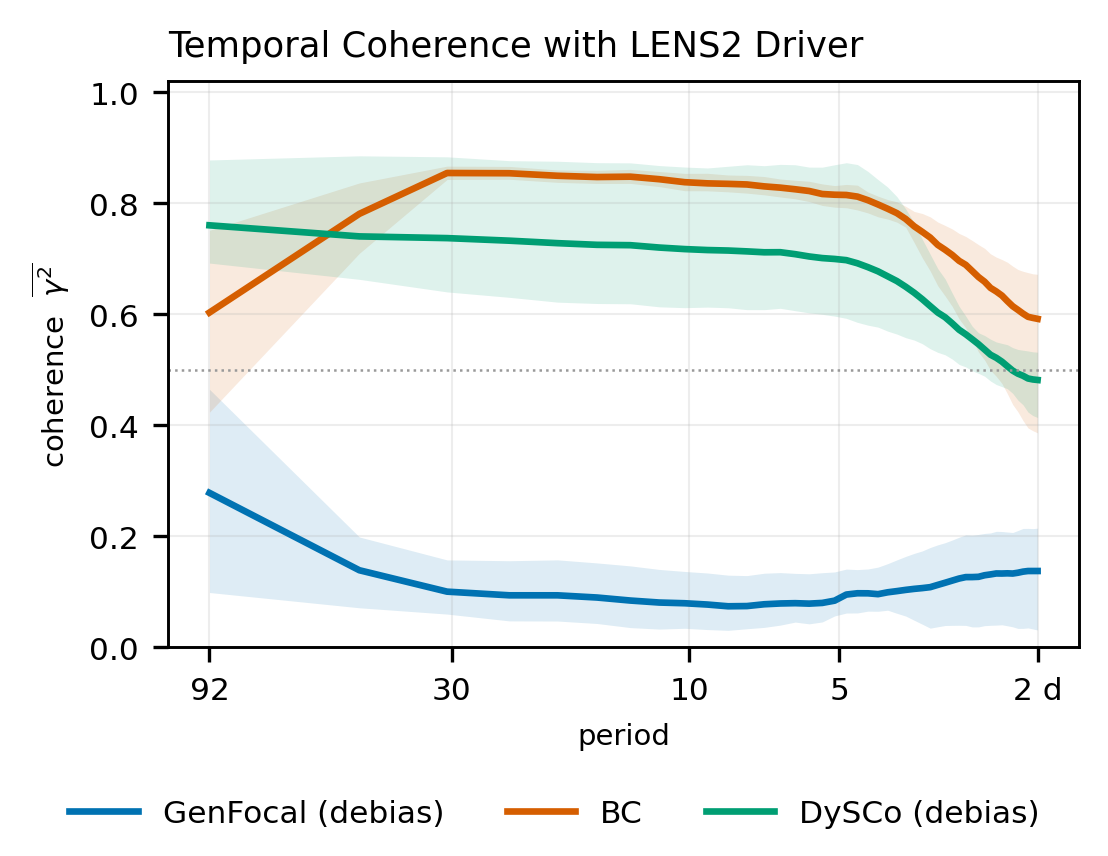}
    \caption{Temporal coherence between input LENS2 driver and debiased fields. Shaded region represents the 5$^{\rm th}$ and 95$^{\rm th}$ percentile range.}
    \label{fig:_427}
\end{figure}

\end{document}